\documentclass[conference]{IEEEtran}
\usepackage{cite}
\usepackage{amsmath,amssymb,amsfonts}
\usepackage{algorithmic}
\usepackage{graphicx}
\usepackage{textcomp}
\usepackage{xcolor}
\usepackage{graphicx}
\usepackage{amsmath}
\usepackage{algorithmic}
\usepackage{latexsym}
\usepackage{makecell}
\usepackage{multirow}
\usepackage{subfig}
\usepackage{dblfloatfix}
\usepackage{float}
\usepackage{colortbl}
\usepackage{tabularx}
\usepackage{longtable}

\usepackage{array}
\usepackage[utf8]{inputenc} 
\usepackage[T1]{fontenc}    
\usepackage[hidelinks]{hyperref}       
\usepackage{url}            
\usepackage{booktabs}       
\usepackage{nicefrac}       
\usepackage{microtype}      
\usepackage{fancyhdr}
\usepackage{lipsum} 

\fancypagestyle{plain}{
  \fancyhf{}
  \fancyfoot[C]{\thepage}%
}
\fancypagestyle{firstpage}{
  \fancyhf{}
  \fancyfoot[C]{\small © 2022 IEEE. Personal use of this material is permitted. Permission from IEEE must be obtained for all other uses, in any current or future media, including reprinting/republishing this material for advertising or promotional purposes, creating new collective works, for resale or redistribution to servers or lists, or reuse of any copyrighted component of this work in other works.}%
}

\def\BibTeX{{\rm B\kern-.05em{\sc i\kern-.025em b}\kern-.08em
    T\kern-.1667em\lower.7ex\hbox{E}\kern-.125emX}}
\begin{document}
\bstctlcite{IEEEexample:BSTcontrol}

\title{A Multimodal Approach for Dementia Detection from Spontaneous Speech with Tensor Fusion Layer\\
}

\author{\IEEEauthorblockN{Loukas Ilias, Dimitris Askounis, and John Psarras}
\IEEEauthorblockA{\textit{Decision Support Systems Laboratory} \\ \textit{School of Electrical and Computer Engineering} \\
\textit{National Technical University of Athens}\\
15780 Athens, Greece \\
\{lilias,askous,john\}@epu.ntua.gr}
}

\maketitle

\begin{abstract}
Alzheimer's disease (AD) is a progressive neurological disorder, meaning that the symptoms develop gradually throughout the years. It is also the main cause of dementia, which affects memory, thinking skills, and mental abilities. Nowadays, researchers have moved their interest towards AD detection from spontaneous speech, since it constitutes a time-effective procedure. However, existing state-of-the-art works proposing multimodal approaches do not take into consideration the inter- and intra-modal interactions and propose early and late fusion approaches. To tackle these limitations, we propose deep neural networks, which can be trained in an end-to-end trainable way and capture the inter- and intra-modal interactions. Firstly, each audio file is converted to an image consisting of three channels, i.e., log-Mel spectrogram, delta, and delta-delta. Next, each transcript is passed through a BERT model followed by a gated self-attention layer. Similarly, each image is passed through a Swin Transformer followed by an independent gated self-attention layer. Acoustic features are extracted also from each audio file. Finally, the representation vectors from the different modalities are fed to a tensor fusion layer for capturing the inter-modal interactions. Extensive experiments conducted on the ADReSS Challenge dataset indicate that our introduced approaches obtain valuable advantages over existing research initiatives reaching Accuracy and F1-score up to 86.25\% and 85.48\% respectively.
\end{abstract}


\section{Introduction}
\thispagestyle{firstpage}

Alzheimer's disease (AD) constitutes a progressive brain disorder, which is ranked as the seventh leading cause of death in the United States and is the main cause of dementia among older adults \cite{alzheimers_disease}. Dementia comes with a group of symptoms and affects memory, behaviour, thinking and social abilities \cite{gauthier2021world}. Nowadays, researchers use spontaneous speech for detecting AD patients, since the detection of AD patients through Magnetic Resonance Imaging (MRI), Positron Emission Tomography (PET), etc. requires access to medical centers, and demands time and money. Recently, there have been proposed shared tasks \cite{luz20_interspeech,luz21_interspeech}, where the researchers can introduce models for detecting dementia from spontaneous speech.

Although there have been proposed several studies, which adopt multimodal approaches by exploiting both speech and transcripts, there are still substantial limitations. More specifically, existing research initiatives add or concatenate the representation vectors obtained by different modalities \cite{10.3389/fcomp.2021.624683} during training, treating equally each modality and consequently obtaining suboptimal performance. In addition, research works employ early \cite{pompili20_interspeech, martinc20_interspeech} and late fusion approaches \cite{mittal2021multimodal,cummins2020comparison, pappagari20_interspeech, sarawgi20_interspeech,syed20_interspeech}. Employing early fusion approaches means that the representation vectors of the different modalities are concatenated at the input level, while exploiting late fusion approaches means that multiple models are trained for each modality separately and the final result is obtained either by a straightforward majority-vote approach or by combining the results of each classifier through a weighted manner. Except for increasing the training time, none of these approaches captures the inter- and intra-modal interactions.

Motivated by these limitations, we introduce deep neural networks, which are trained in an end-to-end trainable manner and take into account both the inter- and intra-modal interactions. First, we convert each audio file into a log-Mel spectrogram, its delta, and delta-delta. Thus, we create an image consisting of three channels. We pass each transcript through a BERT model followed by a self-attention mechanism incorporating a gating model for capturing the intra-modal interactions. Similarly, we pass each image through a Swin Transformer followed by an independent self-attention mechanism with a gate model. We extract also acoustic features from the audio files. Next, the representation vectors from the three different modalities are passed to a tensor fusion layer, which captures effectively the inter-modal interactions. The output of the tensor fusion layer is passed through a series of dense layers for classifying the subject into an AD patient or a non-AD one.

Our main contributions can be summarized as follows:

\begin{itemize}
    \item To the best of our knowledge, this is the first study proposing a Swin Transformer in the task of dementia detection from spontaneous speech.
    \item We introduce the tensor fusion layer, which can capture the inter-modal interactions. There is no prior work utilizing a tensor fusion layer for the task of dementia detection from spontaneous speech.
    \item Our proposed architectures achieve comparable results to existing state-of-the-art approaches.
\end{itemize}

\section{Task and Data}

Given a labelled dataset consisting of AD and non-AD patients, the task is to identify if an audio file along with its transcript belongs to an AD patient or to a non-AD one.

We use the ADReSS Challenge dataset for conducting our experiments
\cite{luz20_interspeech}. This dataset comprises speech recordings and transcripts of spoken picture descriptions elicited from participants through the Cookie Theft picture from the Boston Diagnostic Aphasia Exam \cite{10.1001/archneur.1994.00540180063015}. We opted for this dataset, since it minimizes biases often overlooked in evaluations of AD detection methods, including repeated occurrences of speech from the same participant, variations in audio quality, and imbalances of gender and age distribution. It consists of a train and a test set. The train set consists of 54 AD and 54 non-AD patients, while the test set includes 24 AD and 24 non-AD patients. One further advantage of using the ADReSS Challenge dataset is pertinent to its speaker independent nature.

\section{Predictive Models}
\noindent \textbf{BERT+Gated Self-Attention:} We use the Python library \textit{PyLangAcq} \cite{lee-et-al-pylangacq:2016} for having access to the transcripts, since the manual transcripts have been annotated using the CHAT coding system \cite{10.1162/coli.2000.26.4.657}. We pass each transcript through a BERT model \cite{devlin-etal-2019-bert}. Let $Z \in \mathcal{R}^{N \times d}$ be the output of the BERT model. After this, we add positional encodings to the output of BERT \cite{10.5555/3295222.3295349}. Next, we pass $Z$ through a gated self-attention mechanism \cite{yu2019multimodal}, as described by the equations below:

\begin{equation}
    Q = Z, K = Z, V = Z
    \label{first_equation}
\end{equation}

Next, we adopt the gating model introduced by \cite{yu2019multimodal} as follows:
\begin{equation}
    M = \sigma \left (FC^g \left(FC^g _q \left(Q\right)\odot FC^g _k \left(K\right)\right)\right )
    \label{gating_model}
\end{equation}
where $FC^g _q, FC^g _k \in \mathbb{R}^{d \times d_g}$, $FC^g \in \mathbb{R}^{d_g \times 2}$ are three fully-connected layers, and $d_g$ denotes the dimensionality of the projected space corresponding to 64 units. $\odot$ denotes the element-wise product function and $\sigma$ the sigmoid function. In addition, $M \in \mathbb{R}^{N \times 2}$ corresponds to the two masks $M_q \in \mathbb{R}^N$ and $M_k \in \mathbb{R}^N$ for the features $Q$ and $K$ respectively.

Next, the two masks $M$ and $K$ are tiled to $\Tilde{M_q}, \Tilde{M_k} \in \mathbb{R}^{N \times d}$ and then used for computing the attention map as following:
\begin{equation}
    A^g = softmax \left (\frac{\left(Q \odot \Tilde{M_q}\right)\left(K \odot \Tilde{M_k}\right)^T}{\sqrt{d}} \right )
    \label{eq_gating_model}
\end{equation}

\begin{equation}
    H = A^g V
    \label{a_g}
\end{equation}

Next, we pass $H$ through an average pooling layer followed by a dense layer with 128 units. Let $z^t \in \mathcal{R}^{128}$ be the representation vector corresponding to the transcripts, i.e., textual modality.

\noindent \textbf{Swin Transformer+Gated Self-Attention:} We use the Python library \textit{librosa} \cite{brian_mcfee_2022_6097378} and convert each audio file into an image consisting of three channels, namely log-Mel spectrogram, its delta, and its delta-delta. For all the experiments conducted, we use 224 Mel bands, hop length equal to 1024, and a Hanning window. Each image is resized to $\left (224 \times 224 \right)$ pixels. Next, each image is passed through a Swin Transformer \cite{Liu_2021_ICCV}. Let $Z \in \mathcal{R}^{T \times d}$ be the output of the Swin Transformer. After this, we add positional encodings to the outputs of the Swin Transformer. Next, we pass $Z$ through an independent gated self-attention, as described via the equations \ref{first_equation}-\ref{a_g}. The output of the gated self-attention mechanism is passed through an average pooling layer followed by a dense layer consisting of 32 units. Let $z^v \in \mathcal{R}^{32}$ be the representation vector corresponding to the visual modality.

\noindent \textbf{eGeMAPS:} We use the openSMILE Toolkit \cite{10.1145/1873951.1874246} and extract the acoustic feature set, namely eGeMAPSv02 (functionals), per audio file. In this way, each feature set per audio file has a dimensionality of 88d. Then, we use a dense layer and project the dimensionality to 32. Let $z^{\alpha} \in \mathcal{R}^{32}$ be the representation vector for the acoustic modality.

\noindent \textbf{Tensor Fusion Layer:} We pass $z^t, z^v, z^{\alpha}$ through a tensor fusion layer \cite{zadeh-etal-2017-tensor} for capturing the inter-modal interactions, as described via the equations below.

\begin{equation}
    \left \{ \left(z^t, z^v, z^{\alpha} \right) | z^t \in \begin{bmatrix} z^t \\1 \end{bmatrix}, z^v \in \begin{bmatrix} z^v \\1 \end{bmatrix}, z^{\alpha} \in \begin{bmatrix} z^{\alpha} \\1 \end{bmatrix} \right \}
\end{equation}

\begin{equation}
    z^m = \begin{bmatrix} z^t \\1 \end{bmatrix} \otimes \begin{bmatrix} z^v \\1 \end{bmatrix} \otimes \begin{bmatrix} z^{\alpha} \\1 \end{bmatrix}
\end{equation}
, where $\otimes$ indicates the outer product between the vectors. Let $z^m \in \mathcal{R}^{129 \times 33 \times 33}$ be the output of the tensor fusion layer.

\noindent \textbf{Output Layer:} We pass $z^m$ to a dropout layer with a rate of 0.6 followed by a dense layer of 128 units with a ReLU activation function. We use a dropout layer with a rate of 0.2. Finally, we use a dense layer consisting of two units.

\begin{figure*}[]
\centering
\includegraphics[width=1\textwidth]{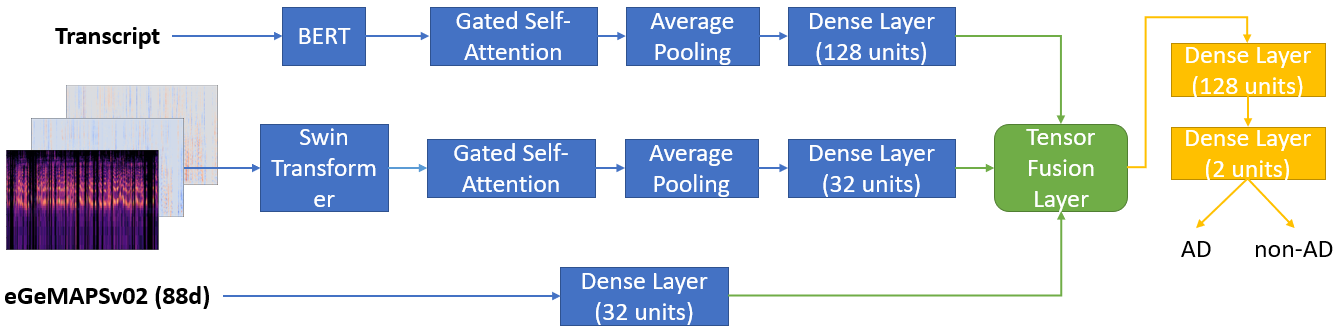}
\caption{Our introduced architecture}
\label{proposed_architecture}
\end{figure*}

The proposed architecture (\textbf{Transcript+Image+Acoustic}) is illustrated in Fig.~\ref{proposed_architecture}.\footnote{We experiment with multiple inputs, namely \textbf{Transcript+Acoustic} and \textbf{Transcript+Image}.}

\section{Experiments}

\subsection{Comparison with state-of-the-art approaches}

We compare our introduced architectures with the following research works, which have proposed multimodal approaches and have reported the results on the ADReSS test set: \textbf{(1)} Fusion Maj. (3-best) \cite{cummins2020comparison}, \textbf{(2)} System 3: Phonemes and Audio \cite{edwards20_interspeech}, \textbf{(3)} Fusion of system \cite{pompili20_interspeech}, \textbf{(4)} Bimodal Network (Ensembled Output) \cite{koo20_interspeech}, \textbf{(5)} GFI, NUW, Duration, Character 4-grams, Suffixes, POS tag, UD \cite{martinc20_interspeech}, \textbf{(6)} Acoustic \& Transcript \cite{pappagari20_interspeech}, \textbf{(7)} Dual BERT (Concat/Joint, BERT large) \cite{10.3389/fcomp.2021.624683}, \textbf{(8)} Model C \cite{10.3389/fnagi.2021.623607}, \textbf{(9)} Majority vote (NLP + Acoustic) \cite{10.3389/fcomp.2021.624659}, \textbf{(10)} LSTM with Gating (Acoustic + Lexical + Dis) \cite{rohanian20_interspeech}, \textbf{(11)} Ensemble \cite{sarawgi20_interspeech}, and \textbf{(12)} Attempt 4 \cite{syed20_interspeech}.

\subsection{Experimental Setup}

We follow a similar training strategy to the one adopted by \cite{10.3389/fcomp.2021.624683}. Firstly, we divide the train set provided by the Challenge into a train and a validation set (65\%-35\%). Next, we train the proposed architectures five times with an Adam optimizer and a learning rate of 1e-5. We minimize the cross-entropy loss function. We apply \textit{ReduceLROnPlateau}, where we reduce the learning rate by a factor of 0.1, if the validation loss has stopped decreasing for three consecutive epochs. Also, we apply \textit{EarlyStopping} and stop training, if the validation loss has stopped decreasing for six consecutive epochs. We test the proposed models using the test set provided by the Challenge. All models are created using the PyTorch library \cite{NEURIPS2019_9015}. We use the Swin Transformer\footnote{microsoft/swin-tiny-patch4-window7-224} and the BERT base uncased version from the Transformers library \cite{wolf-etal-2020-transformers}. All experiments are conducted on a single Tesla P100-PCIE-16GB GPU.

\subsection{Evaluation Metrics}

Accuracy, Precision, Recall, F1-Score, and Specificity have been used for evaluating the results of the introduced architectures. These metrics have been computed by regarding the dementia class as the positive one. We report the average and standard deviation of these metrics over five runs.

\section{Results}
Table \ref{Comparison_With_State_of_the_Art_Approaches} reports the results of our introduced architectures. In this table, we also compare the results of our approaches with the multimodal state-of-the-art ones.

Regarding our proposed models, one can observe that \textit{Transcript+Image+Acoustic} constitutes our best performing model in terms of Precision, F1-score, Accuracy, and Specificity. Although there are models surpassing \textit{Transcript+Image+Acoustic} in Recall, \textit{Transcript+Image+Acoustic} surpasses them in F1-score, which is the weighted average of Precision and Recall. In addition, F1-score is a more important metric than Specificity in health-related problems, since high Specificity and low F1-score means that AD patients are misdiagnosed as non-AD ones. \textit{Transcript+Image+Acoustic} obtains an Accuracy and F1-score of 86.25\% and 85.48\% respectively. It outperforms the other introduced models in Accuracy by 1.67-3.75\%, in F1-score by 0.40-2.14\%, in Precision by 7.37-11.29\%, and in Specificity by 10.00-14.16\%.

In comparison with the existing research initiatives, one can observe that our best performing model, namely \textit{Transcript+Image+Acoustic}, outperforms the other approaches in terms of Accuracy and F1-score. Specifically, \textit{Transcript+Image+Acoustic} surpasses the research initiatives in Accuracy by 1.05-13.33\%. Also, it surpasses the state-of-the-art approaches in F1-score by 0.08-15.72\%. With regards to \textit{Transcript+Image}, it surpasses the research initiatives, except \cite{cummins2020comparison}, in Accuracy by 1.58-11.66\% and in F1-score by 2.08-15.32\%. In terms of \textit{Transcript+Acoustic}, it outperforms the multimodal state-of-the-art approaches, except \cite{cummins2020comparison,10.3389/fcomp.2021.624683,10.3389/fcomp.2021.624659,sarawgi20_interspeech}, in Accuracy by 1.25-9.58\%, while it outperforms the research initiatives, except \cite{cummins2020comparison}, in F1-score by 0.34-13.58\%.

Therefore, one can observe that our introduced advanced transformer-based networks along with techniques capturing the inter-modal interactions obtain comparable or better performance than existing research initiatives, while they require less time for training.

\begin{table}[hbt]
\centering
\caption{Performance comparison among proposed models and state-of-the-art approaches on the ADReSS Challenge test set. Reported values are mean $\pm$ standard deviation. Results are averaged across five runs.}
\begin{tabular}{lccccc}
\toprule
\multicolumn{1}{c}{} & \multicolumn{5}{c}{\small \textbf{Evaluation Metrics}} \\
\cline{2-6}
\multicolumn{1}{l}{\textbf{\small Architecture}}&\textbf{\small Prec}&\textbf{\small Rec}&\textbf{\small F1-score}&\textbf{\small Acc}&\textbf{\small Spec}\\
\midrule
\multicolumn{6}{>{\columncolor[gray]{.8}}l}{\textbf{\small Multimodal state-of-the-art approaches}} \\
\small \textit{\makecell[l]{\cite{cummins2020comparison}}} & \small - & \small- & \small85.40 & \small85.20 &\small -\\
\hline
\small \textit{\makecell[l]{\cite{edwards20_interspeech}}} & \small81.82 & \small75.00 & \small78.26 & \small79.17 & \small83.33\\
\hline
\small \textit{\cite{pompili20_interspeech}} & \small 94.12 & \small66.67 & \small78.05 & \small81.25 &\small 95.83 \\
\hline
\small \textit{\makecell[l]{\cite{koo20_interspeech}}} & \small89.47 & \small70.83 & \small79.07 & \small81.25 & \small 91.67\\
\hline
\small \textit{\makecell[l]{\cite{martinc20_interspeech}}} & \small- &\small - & \small- &\small77.08 & \small-\\
\hline
\small \textit{\makecell[l]{\cite{pappagari20_interspeech}}} & \small70.00 & \small88.00 & \small78.00 & \small75.00 & \small 63.00\\
\hline
\small \textit{\cite{10.3389/fcomp.2021.624683}} & \small83.04 & \small83.33 & \small82.92 & \small82.92 & \small 82.50\\
& \small $\pm$3.97 & \small $\pm$5.89 & \small $\pm$1.86 & \small $\pm$1.56 & \small $\pm$5.53\\
\hline
\small \textit{\cite{10.3389/fnagi.2021.623607}} & \small78.94 & \small62.50 &\small 69.76 & \small72.92 & \small83.33\\
\hline
\small \textit{\makecell[l]{\cite{10.3389/fcomp.2021.624659}}} & \small- & \small- & \small- & \small83.00 & \small-\\
\hline
\small \textit{\makecell[l]{\cite{rohanian20_interspeech}}} & \small81.82 & \small75.00 & \small78.26 & \small79.17 & \small83.33 \\ \hline
\small \textit{\cite{sarawgi20_interspeech}} & \small83.00 & \small83.00 & \small83.00 & \small83.00 &\small -\\ \hline
\small \textit{\cite{syed20_interspeech}} & \small- & \small- & \small- & \small79.17 &\small -\\ 
\midrule
\multicolumn{6}{>{\columncolor[gray]{.8}}l}{\textbf{\small Our introduced models}} \\
\small \textit{\small{Transcript+}} & \small 79.59 & \small 87.50 & \small 83.34 & \small 82.50 & \small 77.50 \\ 
\small \textit{\small{Acoustic}}& \small $\pm$2.66 & \small $\pm$2.64 & \small $\pm$2.35 & \small $\pm$2.49 & \small $\pm$3.33 \\ \hline
\small \textit{\small{Transcript+}} & \small83.51 & \small87.50 & \small85.08 & \small84.58 & \small81.66 \\ 
\small \textit{\small{Image}}& \small $\pm$4.25 & \small $\pm$3.73 & \small $\pm$1.84 & \small $\pm$2.49 & \small $\pm$7.73 \\ \hline
\small \textit{\small{Transcript+}} & \small90.88 &\small 80.83 & \small85.48 & \small86.25 & \small91.66 \\
\small \textit{\small{Image+Acoustic}}& \small $\pm$3.60 & \small $\pm$2.04 & \small $\pm$0.76 & \small $\pm$1.02 & \small $\pm$3.73 \\ 
\bottomrule
\end{tabular}
\label{Comparison_With_State_of_the_Art_Approaches}
\end{table}

\section{Limitations}

\noindent \textbf{GPU resources - Hyperparameter Tuning:} Due to limited access to GPU resources, we did not perform hyperparameter tuning. To be more precise, the number of units in the dense layers, the number of the dense layers, the learning rate, etc. were not chosen via tuning. Performing hyperparameter tuning often leads to an increase in the classification performance.

\noindent \textbf{Explainability - Misclassifications:} In this study, we did not apply explainability techniques for providing to the reader an error analysis, i.e., find the samples that our introduced model fails to classify correctly. In the future, we aim to apply explainability techniques, such as integrated gradients, for rendering the introduced model explainable. In this way, we will able to understand why our model fails to classify some subjects correctly.

\noindent \textbf{Generalizability:} In this paper, we used one dataset for conducting our experiments. For proving the generalizability of our introduced model, we aim to exploit more datasets in the future.

\section{Conclusion and Future Work}

In this paper, we present the first study for the task of dementia detection from spontaneous speech, which employs a Swin Transformer and a tensor fusion layer for extracting visual information and capturing the inter-modal interactions respectively. Specifically, the representation vectors corresponding to the three modalities, i.e., textual, visual, and acoustic, are passed to a tensor fusion layer, which models the inter-modal interactions. Our model achieves comparable performance to existing research initiatives reaching Accuracy and F1-score up to 86.25\% and 85.48\% respectively.

In the future, we aim to create an application, which will incorporate the introduced model and detect AD patients. In addition, we plan to propose advanced transfer learning techniques, in order to employ our introduced models in other tasks, including the detection of Parkinson's disease. Also, explainability techniques, including integrated gradients, are one of our plans for understanding the reasons of misclassifications.

\bibliographystyle{IEEEtran}
\bibliography{references}


\vspace{12pt}

\end{document}